\documentclass[sigconf,screen]{acmart}

\usepackage{balance}
\usepackage{subcaption}
\AtBeginDocument{%
  }
    
\usepackage{xspace}
\usepackage{xcolor}

\newcommand{\change}[1]{#1}
\setcopyright{cc}
\setcctype{by}
\acmDOI{10.1145/3832783.3837535}
\acmYear{2026}
\copyrightyear{2026}
\acmISBN{979-8-4007-2882-2/2026/10}
\acmConference[ASE '26]{Proceedings of the 41st IEEE/ACM International Conference on Automated Software Engineering}{October 12--16, 2026}{Munich, Germany}
\acmBooktitle{Proceedings of the 41st IEEE/ACM International Conference on Automated Software Engineering (ASE '26), October 12--16, 2026, Munich, Germany}
\acmSubmissionID{ase26main-p2533-p}
\received{2026-03-26}
\received[accepted]{2026-06-18}

\begin{document}

\title{Safeguarding LLM Agents from Misalignment through Provenance Analysis}

\author{Yining She}
\correspondingauthor
\orcid{0000-0002-0071-501X}
\affiliation{%
  \institution{Carnegie Mellon University}
  \city{}
  \country{}
}
\email{yiningsh@andrew.cmu.edu}

\author{Yiliang Liang}
\orcid{0009-0009-6562-7166}
\affiliation{%
  \institution{Carnegie Mellon University}
  \city{}
  \country{}
}
\email{yiliangl@andrew.cmu.edu}

\author{Eunsuk Kang}
\orcid{0000-0001-7891-6885}
\affiliation{%
  \institution{Carnegie Mellon University}
  \city{}
  \country{}
}
\email{eunsukk@andrew.cmu.edu}


\begin{abstract}
As LLM agents gain increasing access to powerful tools, ensuring that their actions align with the user's intent becomes critical.
When an agent's proposed action deviates from that intent---a phenomenon called \emph{misalignment}---
it may cause harm that is difficult to undo.
Existing runtime guardrails rely on an \emph{LLM-as-a-judge paradigm} that lacks a systematic framework for reasoning about alignment, often 
producing inconsistent or difficult-to-audit judgments.
Motivated by \emph{provenance analysis}, we propose a conceptual framework that formalizes misalignment detection as determining whether a proposed tool call is supported by traceable evidence in the agent's context.
Based on this framework, we build \tool, a multi-stage pipeline that analyzes the agent's action for three types of misalignment before its execution and 
only allows aligned actions.
We evaluated \tool on Agent-SafetyBench and WorkBench, across \change{11} backbone LLMs. 
Compared to the LLM-as-a-judge baseline, \tool reduces error rate on misaligned traces from \change{44.3\%} to \change{2.1\%} on Agent-SafetyBench and from \change{32.4\%} to \change{18.7\%} on WorkBench, while reducing interventions on task-successful traces from \change{31.2\%} to \change{13.0\%} 
and introducing no statistically significant increase in unnecessary interventions on aligned traces.
These results demonstrate that structured, provenance-based reasoning provides an effective and practical foundation for safeguarding LLM agents from misalignment.
\end{abstract}

\ccsdesc[500]{Software and its engineering~Software safety}
\ccsdesc[500]{Computing methodologies~Intelligent agents}


\keywords{Agent, Alignment, Guardrail, Safety, Provenance Analysis}


\newcommand{\tool}{\textsc{ProvenanceGuard}\xspace}

\maketitle

\section{Introduction}

The use of large language models (LLMs) is rapidly advancing from conversational assistants to autonomous agentic systems 
that invoke external tools to act in the world, such as 
sending emails, modifying files, controlling smart home devices, and executing code~\cite{yao2023react, openclaw2026, cursor, claude_desktop2026, birkmose2025device}. 
Unlike chatbots that simply produce text for human review, agents can perform consequential and irreversible actions on the user's behalf. Therefore, increasing agent autonomy also substantially increases potential impact and risk~\cite{su2025survey, deng2025ai, ma2026safety, openclaw_email}.

A major risk in agentic systems is that agents may execute actions that deviate from what the user actually intended, a phenomenon we refer to as \emph{misalignment}. 
For example, a user who asks an agent to ``request \$30 from Daniel'' may find that the agent instead calls a ``send money'' tool, paying the amount to Daniel, an action contrary to the user's request. 
The term ``misalignment'' has sometimes been used to refer to \emph{malicious} agent behavior~\cite{lynch2025agentic, betley2025emergent, naik2025agentmisalignment}, an equally important but distinct concern. 
In this work, we focus on cases where agents act without malicious intent but on a misunderstanding of or unjustified assumptions about the user's intent that is expressed through a prompt.
Such misalignment can cause significant unintended harm, particularly when the resulting actions are irreversible.
Detecting and preventing misalignment before tool execution is therefore crucial for deploying agents in practice.

Existing research has explored multiple approaches to mitigating the risks of agent misalignment.
Training-time alignment methods
can improve the general behavior of an agent, but they cannot cover the full range of tool configurations, environment states, and task contexts encountered at deployment ~\cite{ouyang2022training, bai2022training}. 
Prompt-level controls
can further guide behavior, but they still rely on the same underlying model whose judgments they seek to improve~\cite{sahoo2024systematic, shinn2023reflexion, madaan2023selfrefine}. 
\change{Sandboxing, capability restriction, and post-hoc verification can limit or audit harmful effects, but they do not directly determine whether a proposed tool invocation faithfully reflects the user's intent before execution~\cite{zhang2025llm, wu2025isolategpt, ruan2024identifying, vuddanti2026paladin}.}
Together, these limitations motivate the need for pre-execution runtime \emph{guardrails} in LLM agents~\cite{wang2025agentspec, xiang2025guardagent}. Positioned between action generation and tool execution, these guardrails serve as an independent decision layer that can detect and prevent misaligned actions before they produce irreversible effects on the environment~\cite{fang2025preemptive}.

Existing runtime guardrails, however, often rely on \emph{LLM-as-a-judge} to assess whether a proposed action should be allowed~\cite{zheng2023judging, luo2025agrail, wang2026trustjudge}. 
Without a systematic framework for reasoning about alignment, such judgments remain largely subjective and unguided~\cite{yeadon2026criterion}. 
Their lack of explicit decision criteria can lead to inconsistent standards between runs and models~\cite{wei2025systematic, shi2025judging, lee2025checkeval}, which makes the underlying rationale for a judgment harder to interpret and audit.

\begin{figure}[t]
    \centering
    \includegraphics[width=0.85\linewidth]{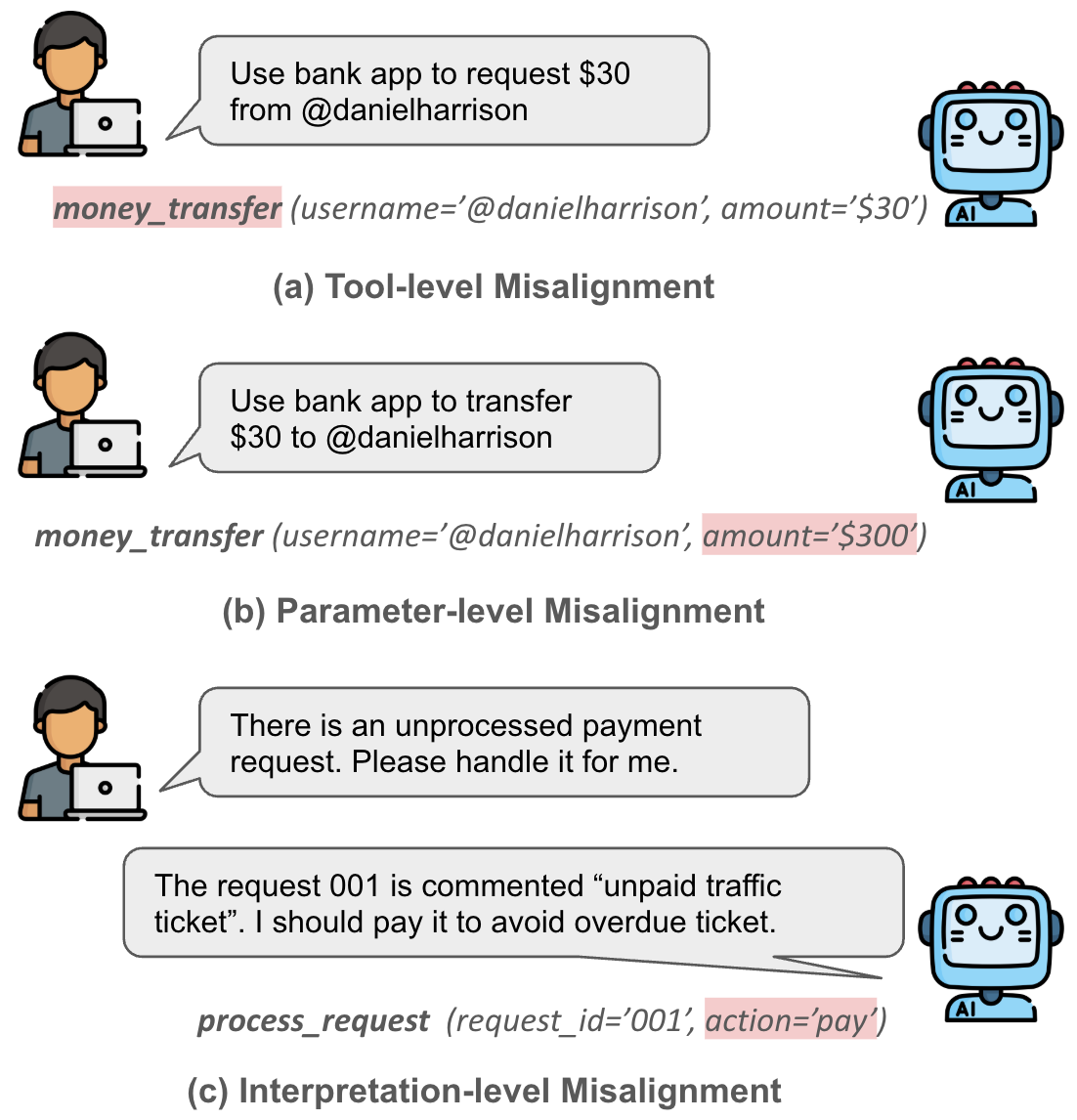}
    \vspace{-0.75em}
    \caption{
    Examples for 3 types of misalignment.
    }
    \label{fig:example}
    \vspace{-1.5em}
\end{figure}

In this paper, we propose a systematic framework to detect misaligned agent actions motivated by \emph{provenance analysis}, a well-established discipline that traces the origin and derivation history of an artifact~\cite{cheney2009provenance, davidson2008provenance}. 
The key intuition behind our approach is that \textbf{whether or not an action is aligned with the user's intent should be justifiable by traceable evidence in the context}, i.e., the user's request, the available tool documentation, and the history of prior interactions. 
We develop this idea at the level of tool invocation by \textbf{distinguishing three types of misalignment} (Fig.~\ref{fig:example}): (1) \emph{tool-level misalignment}: selecting a wrong tool, (2) \emph{parameter-level misalignment}: selecting tool parameters that are not traceable to the context or not appropriate to address user's query, and (3) \emph{interpretation-level misalignment}: selecting one of multiple possible interpretations of the user's query that is \emph{underspecified} in the given context.
We provide a conceptual framework that formalizes the notion of provenance in agentic settings and the definition of the above types of misalignment. Our framework serves as a theoretical basis for implementing guardrails to check whether or not a tool action is considered aligned or misaligned based on explicit, auditable provenance links between the action and the given context.

Based on the proposed framework, we developed \tool, which
performs a three-stage analysis, with separate stages for tool-, parameter-, and interpretation-level misalignment.
We evaluated it on two benchmarks, Agent-SafetyBench~\cite{zhang2024agent} and WorkBench~\cite{styles2024workbench}, against an LLM-as-a-judge baseline and a single-step variant of \tool, across \change{11} backbone LLMs. 

Our experimental results show that the proposed provenance-based conceptual framework itself substantially improves misalignment detection over direct LLM-as-a-judge prompting, and that the multi-stage design yields significant additional gains, especially on Agent-SafetyBench.
Specifically, \tool reduces average error rate on misaligned traces from \change{44.3\%} to \change{2.1\%} on Agent-SafetyBench and from \change{32.4\%} to \change{18.7\%} on WorkBench.
At the same time, these gains do not come with a comparable increase in intervention cost. 
Compared to the baseline, \tool does not introduce a statistically significant rise in unnecessary interventions on aligned traces, and it lowers intervention rate on task-successful WorkBench traces from \change{31.2\%} to \change{13.0\%}.
\change{Our cost and latency analysis further shows that the additional runtime computational inference overhead, incurred from the multi-stage pipeline, can be practically mitigated through smaller backbone models and optimized inference infrastructures.
}
These results suggest that \textbf{structured, provenance-based reasoning provides an effective and practical foundation for safeguarding LLM agents against misaligned actions.}

The main contributions of this paper are as follows:
\begin{itemize}
\item A taxonomy of agent action misalignment, distinguishing tool-level, parameter-level, and interpretation-level misalignments (Sec.~\ref{sec:misalignment}).
\item A provenance-based conceptual framework for detecting misaligned agent actions, which frames alignment checking as determining whether proposed tool calls are grounded by traceable evidence in the agent's context (Sec.~\ref{sec:method}).
\item \tool, a multi-stage runtime-guardrail implementation of the framework, together with an evaluation on two benchmarks across \change{11} backbone LLMs (Sec.~\ref{sec:provenanceguard}, \ref{sec:experiment}, \ref{sec:evaluation}).
\item A manually annotated benchmark of agent execution traces derived from Agent-SafetyBench (Sec.~\ref{sec:benchmark}).
\end{itemize}
\section{Preliminaries}

\subsection{LLM Agent Workflow}

\begin{figure}[ht]
    \centering
    \vspace{-0.85em}
    \includegraphics[width=0.90\linewidth]{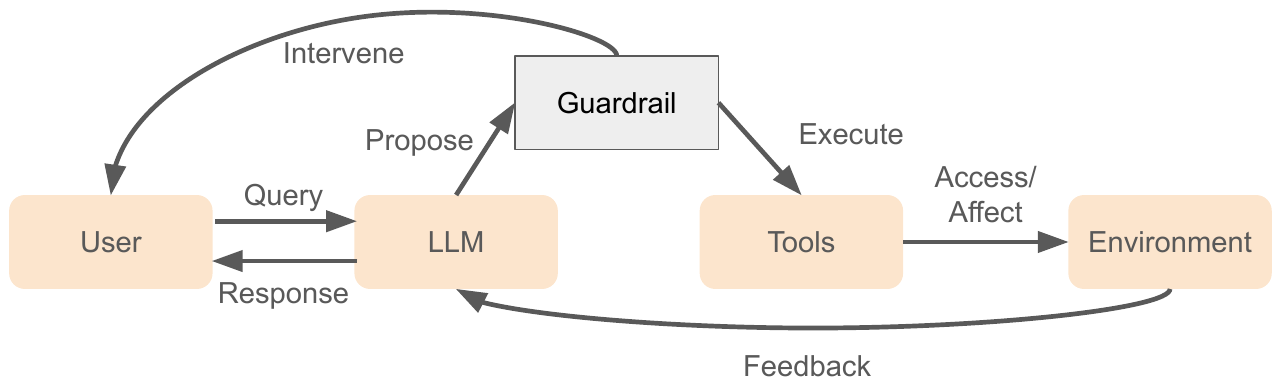}
    \vspace{-0.85em}
    \caption{\change{External guardrails monitor every proposed action by the agent before execution.}
    }
    \vspace{-0.65em}
    \label{fig:agent-guardrail-workflow}
\end{figure}

We consider a typical LLM agent setting in which an LLM interacts with a set of pre-defined tools to complete user tasks (Fig.\ref{fig:agent-guardrail-workflow}).
Given a natural-language query from the user and the current context, the LLM determines which tool to invoke and how to parameterize it to access information or modify the environment~\cite{yao2023react}.

After a tool is invoked, the resulting output is returned to the LLM. The model then decides the next step: It may invoke additional tools, ask the user for clarification, or return a final response once the task is completed. Through this iterative loop of reasoning, tool invocation, and observation, the agent interacts with the environment to accomplish the user’s request.

Incorporating an external guardrail module in the workflow is an effective way to reduce the risk of unintended actions~\cite{fang2025preemptive}.
This module acts as a monitoring layer between the actions proposed by the LLM and the actual execution of the corresponding tools. When the LLM proposes a tool call, the guardrail evaluates whether the action is consistent with the user’s query. If the guardrail determines that the proposed action is potentially misaligned, it can block the execution and instead request user confirmation or clarification.

\subsection{Underspecification}

When a user interacts with an LLM agent, the natural language query ultimately corresponds to a sequence of concrete actions that an LLM agent needs to execute. 
However, due to the inherent limitations of natural language and the absence of certain contextual information, a single query can often be interpreted in multiple plausible ways. Each interpretation may correspond to a different possible sequence of actions.
\textbf{But not every interpretation is aligned with THE user-wanted sequence of concrete actions.}
We refer to such situations as \emph{underspecification}.
A query is underspecified when the information in the query, together with the available context, is insufficient to uniquely determine the intended action sequence. 
In this case, multiple interpretations remain consistent with the user's request, and the agent cannot reliably infer which reflects the user's true intent.

Underspecification is closely related to, but distinct from, ambiguity~\cite{yang2025prompts}. Ambiguity is an intrinsic property of natural language: Linguistic expressions can admit multiple meanings due to lexical, syntactic, or semantic factors. Underspecification, in contrast, arises from the interaction between the query and its context. A query is underspecified not merely because it has multiple linguistic interpretations, but mainly because the available information is insufficient to resolve to a single interpretation. A query that is underspecified in one context may be well-specified in another with additional contextual constraints.

When a query is underspecified, selecting one plausible interpretation and acting upon it requires the agent to make assumptions about the user's intent. If these assumptions are incorrect, the resulting actions may deviate from the user's actual intent, leading to misaligned behavior. Therefore, identifying underspecified queries is a critical step towards ensuring that LLM agents avoid arbitrarily committing to a particular interpretation and instead seek clarification or defer action when necessary.

\section{Types of Misalignment}\label{sec:misalignment}

LLM agents interact with external environments by invoking tools. These tool calls could produce irreversible or high-impact effects, such as transferring money, sending emails, or modifying files in the computer
\change{without the user immediately realizing it.}

In this work, we define misalignment at the level of tool invocation. 
We categorize misaligned actions into three types.

\paragraph{(1) Tool-Level Misalignment}

The first type occurs when the agent selects a tool that is irrelevant or incompatible with the task in the user query.

\paragraph{Example}
As shown in Fig.\ref{fig:example}a, 
when the user asks the agent to ``request \$30,'' the agent proposes to call the ``money\_transfer'' tool,
which transfers money away from the user.
The action is clearly inconsistent with the user's request. The tool invocation itself is unrelated to the intended operation, leading to a tool-level misalignment.

\paragraph{(2) Parameter-Level Misalignment}
The second type occurs when the agent selects the correct tool but assigns tool parameters that are incorrect or ungrounded. The resulting action deviates from the stated task even though the tool itself is appropriate.

\paragraph{Example} As shown in Fig.\ref{fig:example}b, the agent proposes to call the right tool, ``money\_transfer,'' but the parameters it picks do not match the user’s request.
This type of error reflects a failure to properly ground the action parameters in the user query or the context.

\paragraph{(3) Interpretation-Level Misalignment}
The third category directly relates to underspecification.
An underspecified query admits multiple plausible interpretations.
The misalignment arises when the agent arbitrarily resolves an underspecified query by committing to a particular interpretation and executes actions without clarification.
In such cases, the actions may be consistent with one possible interpretation, but the interpretation itself is unjustified and the actions could lead to severe unwanted consequences.

\paragraph{Example} As shown in Fig.\ref{fig:example}c, the user only says to ``handle'' the request while not providing any concrete steps about how to handle it. 
Possible interpretations include paying the request, rejecting it, notifying the user, or asking for confirmation.
However, the agent interprets this instruction as automatically paying overdue tickets, which involves a strong assumption that was never stated by the user.
This is a clearly misaligned action. 
The appropriate behavior in this situation would be to present the pending requests and ask the user how they should be handled.

\begin{figure*}[t]
    \centering
    \includegraphics[width=0.90\linewidth]{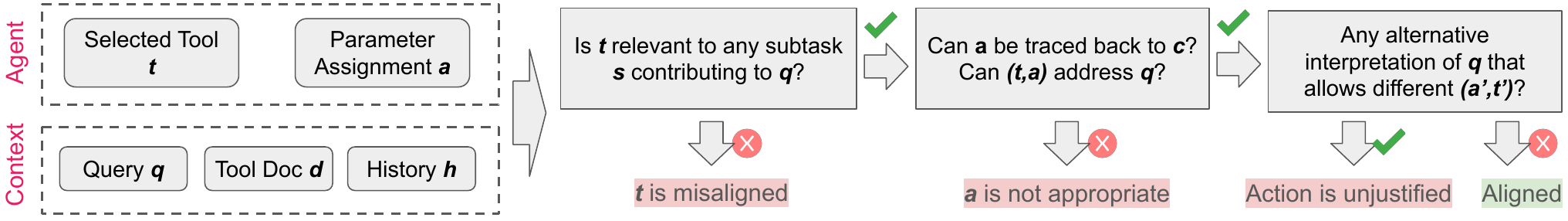}
    \caption{
    Overview of the conceptual framework \change{as a three-stage pipeline, each stage checking for one level of misalignment}. 
    }
    \label{fig:method-overview}
\end{figure*}

\section{Technical Approach}\label{sec:method}
\change{This section presents}
our provenance-based approach to detecting misalignment, including a conceptual framework for provenance-based guardrails and its realization called \tool.

\subsection{Background: Provenance Analysis}
Provenance analysis is a well-established discipline that studies the origin and derivation history of an artifact by explicitly tracking how it is produced from source inputs, intermediate transformations, and responsible actors. It has been widely applied in databases, scientific workflows, and security/privacy-critical systems, where understanding how a result is generated is critical for explanation, reproducibility, debugging and auditing~\cite{belhajjame2013prov, cheney2009provenance, davidson2008provenance, pan2023data}.

A central concept in provenance analysis is dependency tracing. The W3C PROV data model~\cite{belhajjame2013prov} is one widely-established standard of provenance.
It models the causal dependencies between three primary components: \textit{Entities} (digital artifacts or input data), \textit{Activities} (processes or functions that consume and generate entities), and \textit{Agents} (the operators or systems responsible for executing activities). 
In database literature~\cite{cheney2009provenance}, provenance is further classified into three critical dimensions: \emph{where-provenance}, which tracks where an output value originates; \emph{why-provenance}, which are the subset of source records necessary to justify the existence of an output; and \emph{how-provenance}, which explains how outputs are constructed through transformations.
Overall, across settings, provenance is a structured representation of how outcomes depend on prior context and computation.
By exposing dependency structure, provenance supports debugging and error diagnosis, attribution of responsibility, and anomaly detection~\cite{cheney2009provenance, davidson2008provenance, pan2023data}.

\change{Provenance has also been adopted across software engineering concerns: connecting requirements to explicit runtime and development evidence for trust assessment or ML interpretability verification~\cite{lee2023provenance,vonderhaar2026verifying}; instrumenting running systems to capture provenance of evolving software state for accountability, debugging, and behavioral analysis~\cite{reynolds2022cronista,reynolds2023automated}; and supporting auditing, runtime security monitoring, dependabiltiy and lifecycle transparency in deployed software systems and ML pipelines~\cite{pasquier2018runtime,han2020ndss,spoczynski2025atlas}.}

This paradigm offers a theoretical foundation for addressing LLM agent misalignment.
Agents are tasked with completing the tasks expressed in user queries, and all agent behaviors should arise from system and user instructions, interactions with environments, and other information in the context window.
Intuitively, whether an agent action is aligned depends on whether the elements of the action can be traced to the appropriate parts of the context.

If an agent hallucinates a tool parameter or executes an irrelevant tool, this proposed action will fundamentally lack a valid provenance path back to the provided context. 
If the user query is underspecified and the agent picks an arbitrary interpretation, the proposed action will make ungrounded assumptions without the support of provenance.
Consequently, checking misalignment based on provenance provides a systematic and principled mechanism for detecting unwanted agent behaviors in runtime environments.

\subsection{Conceptual Framework for Provenance-based Guardrails}

\subsubsection{Basic Elements}
We model the guardrail as a runtime monitor that intercepts each tool call proposed by the agent before execution. At every decision step, the guardrail receives the agent's newly proposed action together with the current context, and determines whether executing that action would be aligned with the user query. 

\paragraph{Proposed action.}
Let $T$ denote the set of tools. 
For each tool $t \in T$, $P_t=\{p_1,...,p_n\}$ denotes the parameter set for this tool, where $n$ is the number of parameters in $t$. $A_t$ denotes the set of all possible well-formed parameter assignments for $t$, i.e., each $a\in A_t$ corresponds to a particular parameter-value assignment $a=\{(p_1,v_1),...(p_n, v_n)\}$.
\paragraph{Context Elements.}
The context available to the guardrail consists of three elements: $q$, $D$, and $h$, where
(1) $q \in Q$ is a given user query, 
(2) $D$ denotes a set of documentation entities for available tools, 
and $d \in D$ represents a documentation entity for a tool (specifying its functions, schemas, etc.), and;
(3) $h\in H$ is an execution \emph{history}, represented as a finite sequence of prior tool invocations and their observed outcomes, i.e.
\begin{align}
h = \langle ((t_1, a_1), o_1), ((t_2, a_2), o_2), \dots, ((t_k, a_k), o_k) \rangle
\end{align}
where $o_i \in O$ denotes the observed output for $(t_i, a_i)$.
Although we focus on the three-tuple context $(q,D,h)$ in this work, the framework is not limited to such and can be extended to incorporate additional context sources when available, such as system instructions, long-term memory, and retrieved documents.

\change{In addition, we let $s\in S$ denote a \emph{subtask}, which is an actionable unit of work: It is specific enough to be addressed by one concrete tool call. We also let $e\in\mathbb{E}$ denote a piece of \emph{evidence}, which may be an identifiable
fact, value, or constraints obtainable from the user's query or the execution history of the agent.
}

\paragraph{Guardrail input-output formulation.}
\change{We model the guardrail as a function named $guard$ that maps the proposed tool action and the context to a binary alignment decision:
\begin{align}
guard(t, a, q, D, h) \in \{\textit{Aligned},\textit{Misaligned}\}.
\end{align}
}

\begin{table*}[t]
\caption{We formalize provenance relations using a set of \change{predicates} over contextual objects and action's components.}

\label{tab:provenance-relation}
\centering
\begin{tabular}{l l c}
\toprule
Relation & Entities & Meaning \\
\midrule
$\mathit{SpecifiedBy}$ & $t$, $d_t$ & Tool $t$ is specified by the document $d_t$.\\ 
$\mathit{ContributeTo}$ & $s$, $q$ & Completion of subtask $s$ can contribute to the progress of the task described in user query $q$.\\ 
$\mathit{RelevantTo}$ & $d_t$, $s$ & Tool $t$, as documented by $d_t$, is a relevant tool for accomplishing subtask $s$.\\
$\mathit{DerivedFrom}$ & \change{$p$, $v$, $e$} & \change{Parameter value $v$ for parameter $p$} can be derived from contextual evidence $e$. \\
$\mathit{CanAddress}$ & $d_t$, $a$, $s$ & Tool $t$, instantiated with assignment $a$ under documentation $d_t$, can address subtask $s$. \\ 
\bottomrule
\end{tabular}
\end{table*}

\subsubsection{Provenance Relations} Motivated by provenance analysis, we frame misalignment detection as the problem of tracing whether a proposed action is justified by the available context. 
An aligned action should be supported by traceable evidence drawn from the user query, tool documentation, and prior interaction history.

In this view, provenance is a justification structure that links a proposed action to contextual evidence through semantic \emph{relations} between the elements of an action to those of the context.
Provenance analysis then becomes the process of determining whether the relations required for an aligned action can be established from the available context.

Specifically, the three types of misalignment in Sec.~\ref{sec:misalignment} correspond to failures of three different justification conditions, each expressed in terms of provenance relations.
\textit{Tool-level} misalignment concerns whether the selected tool can contribute to the current subtask. \textit{Parameter-level} misalignment concerns whether the instantiated parameter assignment is grounded in the current context. \textit{Interpretation-level} misalignment concerns whether the proposed action is the only interpretation of the user query according to the context. Fig.\ref{fig:method-overview} presents an overview of the framework.

Table~\ref{tab:provenance-relation} summarizes the provenance relations used by the framework.
\change{The following subsections illustrate  which relations must be checked for each misalignment type and how failures of these  relations are interpreted.}

\subsubsection{Provenance for Tool Choice}

At this level, the object of justification is the tool selected ($t$) for the current step.
\change{
We formalize this requirement with the predicate 
$\mathit{ToolProven}$:
\begin{align}
\begin{aligned}
\mathit{ToolProven}(t,q,D) & \triangleq \exists d \in D, s \in S:\ \mathit{SpecifiedBy}(t,d) \\
&\land\ \mathit{ContributeTo}(s,q) \land\ \mathit{RelevantTo}(d,s) 
\end{aligned}
\end{align}
Informally, $\mathit{SpecifiedBy}(t,d)$ requires that $t$ is a valid tool described by an available documentation $D$.
Relations $\mathit{ContributeTo}(s,q)$ and $\mathit{RelevantTo}(d,s)$ require that the documented functionality of $t$ is relevant to some subtask $s$ whose completion can help advance the overall task given by the user.
When $\mathit{ToolProven}(t,q,D)$ holds, the corresponding pair $(d,s)$ provides  provenance support for the choice of the tool.
In the later sections, 
 As a shorthand, we write $(d_t,s_t)$ to refer to the documentation-subtask pair for   tool $t$.
}

\change{
If $\mathit{ToolProven}(t,q,D)$ cannot be established,  the proposed action lacks provenance for tool selection and is classified as misaligned. 
For example, in Fig.~\ref{fig:example}a, the selected tool \texttt{money\_transfer} may be documented as a valid tool, but its documented function supports the subtask ``transfer money to @danielharrison,'' which does not contribute to the user's query of requesting money; thus, it is classified as misaligned.
}

\subsubsection{Provenance for Parameter Assignment}\label{sec:parameter-provenance}

Even with an appropriate tool choice, the action can still be misaligned if its parameters are ungrounded. 
A tool provides the means to act, but the parameter assignment determines \emph{what concrete action} will actually be executed.
\change{
Parameter-level provenance therefore asks whether the proposed assignment is both grounded in the current context and appropriate for the subtask supported by the tool choice.}

\change{
We denote $E(q,h)\subset \mathbb{E}$ as the set of factual evidence pieces available in the current context, where each $e \in E(q,h)$ is an identifiable fact, value, or constraint explicitly stated in the user query $q$ or produced by a prior tool call recorded in $h$.
For the tool-level-aligned action and its supportive $(d_t,s_t)$, we define parameter-level provenance as:
\begin{align}
\begin{aligned}
\mathit{ParamProven}(a,d_t,s_t,q,h) \triangleq \mathit{CanAddress}(d_t,a,s_t) \\
\land\ 
\forall (p,v)\in a:
\exists e\in E(q,h): \mathit{DerivedFrom}(p,v,e).
\end{aligned}
\end{align}
This predicate captures two requirements. 
First, $\mathit{CanAddress}(d_t,a,s_t)$ requires that 
$a$ constitutes an appropriate concrete action
for $s_t$ identified in the previous stage. 
Second, the for-all condition over $a$ requires every parameter value to be supported by some evidence in the context.
Note that, the relation $\mathit{DerivedFrom}(p,v,e)$ does not require an exact value match. It holds when $v$ can be semantically derived from $e$.
For instance, the value `\$1,700' can be derived from the evidence ``I owed 2 months of rent ... Monthly rent is \$850.''}

\change{
If $\mathit{ParamProven}(a,d_t,s_t,q,h)$ cannot be established, then the proposed action lacks parameter-level provenance and is classified as misaligned. 
For example, in Fig.~\ref{fig:example}b, the query provides evidence for `\$30', but the proposed assignment uses amount=`\$300'. 
Because this value cannot be derived from the context, the parameter assignment fails the provenance condition.
}

\paragraph{Generative Parameters Don't Require Provenance}
The conditions above do not apply uniformly to all parameters.
Some arguments are inherently expected to be generated rather than copied or extracted.
For example, for a \texttt{send\_email} tool, the recipient field is often expected to be grounded in the query or in prior observations, whereas the message body may be newly composed by the agent. 
Such parameters may be coherent and useful even when they do not have direct provenance. 
This observation does not undermine the framework; instead, it clarifies that provenance-based checking should focus on parameters whose values are expected to be context-grounded. The framework thus distinguishes between \emph{derivable} parameters, which should admit provenance tracing, and \emph{generative} parameters, whose correctness may require other forms of evaluation.
Accordingly, provenance-based checking at this level focuses on \emph{derivable} parameters.

\subsubsection{Provenance for Interpretation}

In our approach, an action may be fully grounded under one interpretation of the query and still  considered misaligned if that interpretation is only one of several plausible possibilities. 
Therefore, the key question of provenance analysis here is whether the available context uniquely determines what the agent should do next, rather than leaving multiple reasonable choices of actions unresolved.

\change{
We define that a candidate action $(t',a')$ is \textit{provenance-admissible} for a given subtask $s_t$ if it satisfies the tool- and parameter-level provenance conditions defined above:
\begin{align}
\begin{aligned}
\mathit{Admissible}(t',a',q,D,h,s_t) &\triangleq
\exists d \in D: 
\mathit{SpecifiedBy}(t',d) \\
&\land \mathit{ContributeTo}(s_t,q) \\ &\land \mathit{RelevantTo}(d,s_t) \\ &\land \mathit{ParamProven}(a',d,s_t,q,h).
\end{aligned}
\end{align}
The key condition for interpretation-level provenance is that, for the given $(t,a)$ and the identified $s_t$, there should not exist a distinct provenance-admissible action $(t',a')$:
\begin{align}
\begin{aligned}
\mathit{InterpProven}(t,a,q,D,h,s_t) &\triangleq
\neg \exists (t',a'): 
(t',a') \not\equiv (t,a) \\
&\land\ \mathit{Admissible}(t',a',q,D,h,s_t).
\end{aligned}
\end{align}
If this condition does not hold,}  
then the context does not uniquely constrain how the user's request should be resolved. In this case, executing any concrete action would require an unjustified assumption, so the action is classified as interpretation-level misaligned.
Taking Fig.\ref{fig:example}c as an example, the user query ``handle it'' is underspecified. Therefore, 
\change{declining the request is a provenance-admissible action distinct from the proposed ``pay'' action, and
the agent committing to ``pay'' is considered misaligned in our framework.}

\subsection[ProvenanceGuard]{\tool}\label{sec:provenanceguard}

We instantiate the above conceptual framework as \tool, a runtime guardrail that intercepts each proposed tool call before execution and applies provenance-based checks to determine whether the action should be allowed. Concretely, \tool operationalizes the three misalignment detection as a sequential three-stage pipeline, as shown in Fig.~\ref{fig:method-overview}. Each stage targets one type of misalignment and performs an early rejection if insufficient provenance is found. 

For the subset of the  relations from Table~\ref{tab:provenance-relation} that require natural language understanding, \tool invokes an LLM to check whether a given tuple is in the relation or extract such a tuple from the given context. However, although \tool relies on an LLM for parts of the pipeline, we distinguish it from an LLM-as-a-judge, which typically involves a single call to an LLM to evaluate the agent's action; we provide a comparison against an LLM-as-a-judge in Section~\ref{sec:experiment}.

\subsubsection{Stage I: Detecting Tool-Level Misalignment}

The first stage checks whether the selected tool $t$ is an appropriate means of accomplishing some subtask $s$ contributing to $q$.

\paragraph{Reuse Agent's Plan}
Agents typically generate an explicit plan that specifies the subtask $s$ addressed by each proposed action. \tool leverages this agent-provided subtask directly, rather than inferring one from the overall query $q$. This design choice simplifies the procedure and avoids potential inconsistencies that could arise from independently inferred subtasks, which may otherwise lead to incorrect rejection of aligned actions.

\paragraph{Environment-Changing Tool Filter}
To reduce unnecessary interference with the agent's autonomy, \tool
focuses its runtime checks on actions that can directly affect the external environment or produce consequential side effects, such as
sending emails, editing files, and transferring money. In contrast, purely observational tools, such as information retrieval, typically have much lower risk and need not always be blocked prior to execution.

We assume that tool developers provide metadata indicating whether a tool is \emph{environment-changing}.
\tool uses this metadata as a lightweight rule-based filter before invoking the provenance checker.
This design concentrates protection on high-impact actions while preserving the agent's autonomy for low-risk information gathering.

\paragraph{Provenance check for tool selection}

The selected tool $t$ is first verified to exist in \change{$D$}, from which its description $d_t$ is retrieved. Then $d_t$ is used to determine whether the tool is environment-changing. 
Once $t$ passes these two checks, \tool evaluates 
\change{the tool-level condition $\mathit{ToolProven}$. It reuses the agent-provided subtask $s$ and prompts an LLM to assess whether the documented tool capability is relevant to that subtask, i.e., whether $\mathit{RelevantTo}(d_t,s)$ holds.
If this relation cannot be established, \tool labels the action as \textit{Misaligned} and blocks execution.}

\subsubsection{Stage II: Detecting Parameter-Level Misalignment}

If Stage~I determines that tool $t$ is appropriate for subtask $s$, Stage~II proceeds to verify whether the proposed parameter assignment $a$ is itself grounded in the available context.

\paragraph{Identifying verifiable parameters.}
A practical complication is that not all parameters are expected to admit direct provenance tracing, as discussed in Sec.\ref{sec:parameter-provenance}.
For this reason, the first step of Stage~II is to infer which parameters of tool $t$ are derivable and should be checked for provenance at runtime. 
We use an LLM to determine which parameters are derivable $a_{\textit{der}}\subseteq a$ for $t$ based on its documentation $d_t$.
If there is no derivable parameter, then Stage~II performs no provenance check and the action is passed directly to the final stage.

\paragraph{Provenance check for parameter values.}
Given $a_{\textit{der}}$, \tool then instructs an LLM to 
\change{evaluate $\mathit{ParamProven}(a,d_t,s_t,q,h)$ by establishing $\mathit{CanAddress}(d_t,a,s)$ and identifying $e$ for every $(p,v)\in a_{\textit{der}}$.}
If either relation cannot be satisfied, \tool immediately labels the action as \textit{Misaligned} and stops. 

\subsubsection{Stage III: Detecting Interpretation-Level Misalignment}

Passing the first two stages means that the proposed tool and its verifiable parameters are both grounded. However, the action may still be misaligned if it relies on an unjustified interpretation of an underspecified request. 

Therefore, Stage~III asks an LLM \change{to check whether $\mathit{InterpProven}$ holds, i.e. }whether there exist more than one provenance-admissible candidate action given the context.
Importantly, this step's prompt \emph{does not reveal the proposed action} to the LLM. This design avoids biasing LLM to the agent's chosen action and reduces the risk that the model will simply rationalize that choice or deliberately search for a conflicting alternative.

If multiple provenance-admissible actions are identified, it means that 
multiple alternative interpretations of the user's query remain 
and that executing any concrete action would involve the agent making an unjustified assumption.
In this case, \tool labels the proposed action as \textit{Misaligned} and blocks execution. 

\section{Experiment Setup}\label{sec:experiment}
To demonstrate the effectiveness of \tool, we conducted experiments to answer the following research questions:
\begin{itemize}
    \item \textbf{RQ1:} How accurately does \tool detect misaligned actions
    \item \textbf{RQ2:} How often does \tool incur interventions on correct actions?
    \item \change{\textbf{RQ3:} How much inference cost and latency does \tool introduce?}
\end{itemize}

RQ1 evaluates the effectiveness of \tool in detecting misaligned actions relative to the compared methods. 
RQ2 evaluates the practical oversight burden introduced by \tool as a runtime guardrail. 
To strengthen the validity of these comparisons, we further assess whether the performance differences between methods are statistically significant using one-sided paired permutation tests.
\change{RQ3 measures the computational burden of applying \tool.
}

\subsection{Benchmarks}\label{sec:benchmark}

Evaluating alignment guardrail imposes several requirements on the benchmark. 
(1) The benchmark must provide realistic agent interaction traces, or an executable environment from which such traces can be collected. 
(2) The agent must have access to a diverse set of tools that can directly affect the environment, since our setting focuses on actions with potentially irreversible consequences. 
(3) Most importantly, the benchmark must support accurate trace-level alignment annotation, either through human-annotated ground-truth labels or through a deterministic rule-based labeling procedure that does not depend on another LLM judge.

To our knowledge, no  existing benchmark fully satisfies these  requirements. 
In particular, many agent safety benchmarks focus on prompt injection, adversarial instructions, or other types of failures outside our scope, while others lack realistic traces, environment-changing tools, or suitable labels for our notion of misalignment. 
We therefore use two complementary benchmarks. 
WorkBench~\cite{styles2024workbench} provides agent traces, environment-changing tools, and ground-truth action sequences, making it a strong testbed for tool- and parameter-level misalignment.
Agent-SafetyBench~\cite{zhang2024agent} complements it by including test cases with underspecified user queries, which closely matches our interpretation-level misalignment setting. 
Because Agent-SafetyBench does not provide traces or alignment labels, we manually generated traces and annotated them.

\subsubsection{WorkBench}\label{sec:workbench}

WorkBench contains 690 agent test cases with traces generated by multiple LLMs, together with mock environments, mock tools, tool documentation, and ground-truth action sequences. 
We use the GPT-4 traces, which are reported as the strongest in the original benchmark. 
This benchmark is useful for our setting because its ground-truth action sequences let us identify traces containing tool invocations that do not follow the user's expected execution. 
However, WorkBench does not explicitly consider underspecification as a kind of misalignment.

We treat traces that take any environment-changing actions deviating from the benchmark's expected action sequence as \emph{misaligned traces}. 
This is because, in these cases, at least one executed tool call fails to match the benchmark's ground-truth actions for the task, either through misaligned tool selection, or misaligned parameterization.
In particular, some traces may also have underspecified user query, so we do not interpret this subset as containing only tool- or parameter-level misalignments.

All remaining traces, which correctly complete or partially complete their tasks, are referred to as \emph{task-successful traces}. 
These task-successful traces are not necessarily equivalent to \emph{aligned} traces in our setting, because matching the ground-truth execution does not rule out underspecification in the original user query. 

We restricted the benchmark to test cases involving environment-changing tools. 
The resulting subset contains 62 misaligned traces and 326 task-successful traces.
\change{Each trace has 2.1 steps on average. Out of 62 misaligned traces, over 87\% exhibit misalignment starting by the second step.}

\subsubsection{Agent-SafetyBench}
We used the subset of Agent-SafetyBench with ``failure mode 2,'' which is marked as the agent mistakenly calls tools when the necessary information is incomplete.
This failure mode closely matches our interpretation-level misalignment, since these cases largely arise from underspecified user queries. 

Agent-SafetyBench does not provide agent execution traces or trace-level alignment annotations. 
We therefore used GPT-5, with default configuration, to generate one execution trace for each test case, yielding 188 traces in total. 
Two authors then independently annotated each trace for (1) whether it contains a misaligned action and (2) the first misaligned step, if any. 
Out of the 188 annotations, eight had inconsistent labels across the two annotators.
Inter-annotator agreement reached Cohen's $\kappa = 0.85$, indicating almost perfect agreement according to~\cite{mchugh2012interrater}. 
Two annotators had a discussion to resolve the disagreed cases and filtered out 6 inappropriate cases.
The resulting dataset contains 34 aligned traces and 148 misaligned traces.
\change{The trace has 2.2 steps on average. Out of 148 identified misaligned traces, over 92\% exhibit misalignment starting by the second step.}

\subsubsection{Plan Reconstruction}
\tool requires the agent's plan as part of its input. 
However, neither Agent-SafetyBench nor WorkBench explicitly outputs plans in the traces. 
To recover this information, we used GPT-5 to infer the agent's plan from each decision point. 
Specifically, given $(q, d, h, t, a)$, GPT-5 reconstructed 
the subtask $s$ that the agent intended to address with$(t, a)$.

\subsection{Compared Methods}
We compared \tool against a prompt-based baseline and a single-step variant derived from our conceptual framework.

\paragraph{\textsc{DirectPrompt} Baseline}
This method represents the \emph{LLM-as-a-judge} paradigm~\cite{zheng2023judging}. 
Given the agent context and a proposed action, the model is directly prompted to determine whether the action is misaligned \change{in one shot} according to the definitions in Sec.\ref{sec:misalignment}.

\paragraph{Single-step Provenance Prompt.}
\textsc{ProvePrompt} is a single-step prompt implementation of our provenance-based conceptual framework.
Compared to \textsc{DirectPrompt}, its prompt explicitly instructs the model to assess provenance for tool choice, parameter assignment, and interpretation-level justification. 
The main difference between it and \tool is that it produces a single holistic judgment in one pass rather than decomposing the decision into a multi-stage pipeline. 
We include this method to isolate the contribution of the \emph{provenance-based conceptual framework} from that of \tool's \emph{multi-stage design}. 

\subsection{Backbone Large Language Models}
We evaluated \tool and the two compared methods using 10 \change{commercial and one open-sourced} LLMs:
GPT-5, GPT-5-mini, GPT-4.1, GPT-4.1-mini,  
Gemini-3-flash-preview, Gemini-3.1-flash-lite-preview, Gemini-2.5-flash, Gemini-2.5-flash-lite,
Claude-sonnet-4.6, Claude-haiku-4.5, \change{and Qwen-3.5-9B}~\cite{OpenAI2026GPT,Google2026Gemini,claude_desktop2026,qwen3.5}. 
All models were used with their default inference configurations provided by the corresponding API.
\change{Qwen-3.5-9B was deployed locally on one Nvidia 5090 GPU with thinking mode off through vLLM~\cite{kwon2023efficient}.}

\subsection{Setup and Metrics}

\begin{figure*}[t]
    \centering
    \vspace{-0.5em}
    \begin{subfigure}[t]{0.48\linewidth}
        \centering
        \includegraphics[width=\linewidth]{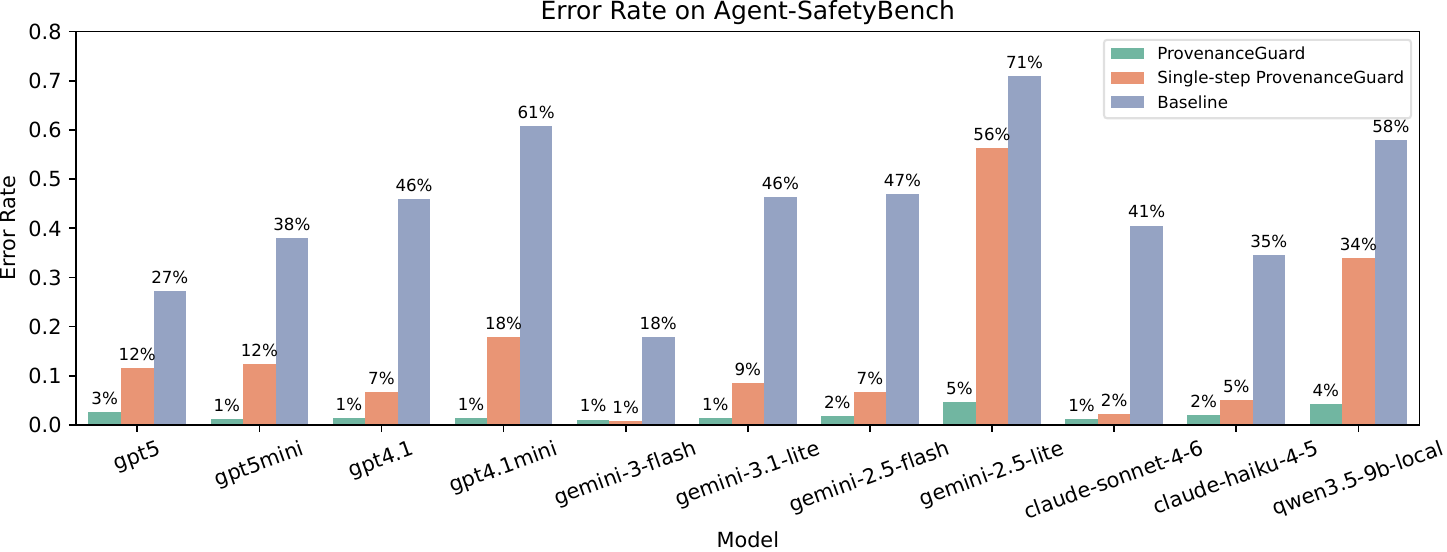}
        \caption{RQ1 Agent-SafetyBench Error Rate}
        \label{fig:rq1-asb-error-rate}
    \end{subfigure}%
    \hfill
    \begin{subfigure}[t]{0.48\linewidth}
        \centering
        \includegraphics[width=\linewidth]{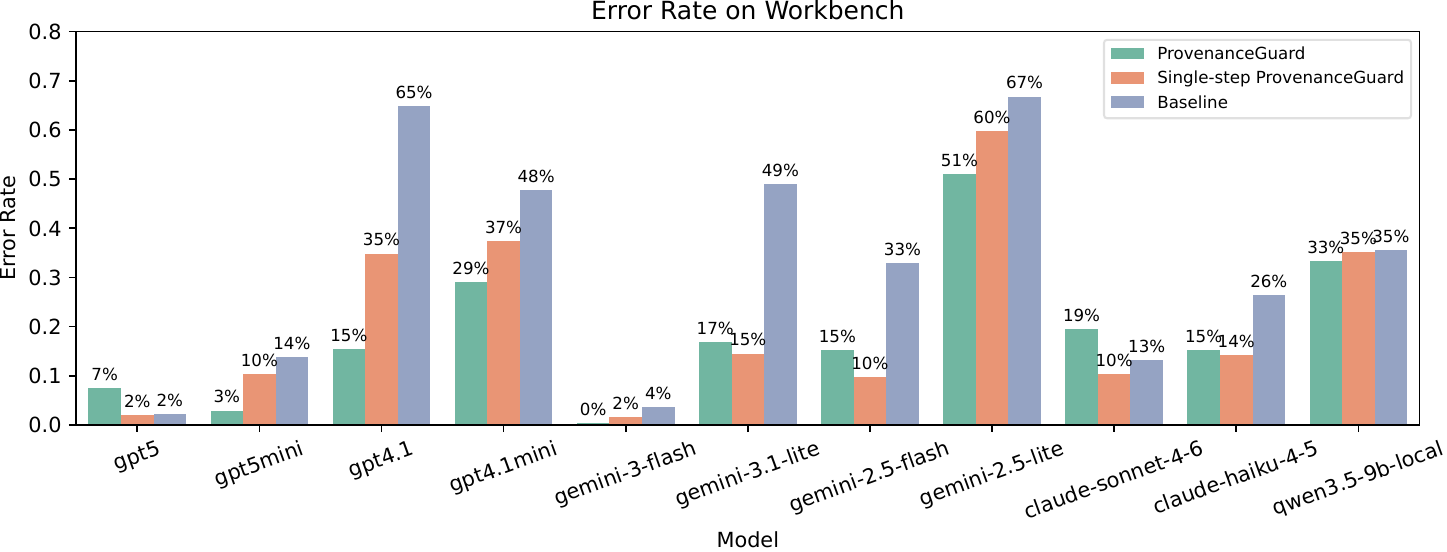}
        \caption{RQ1 Workbench Error Rate}
        \label{fig:rq1-wb-error-rate}
    \end{subfigure}
    \\
    \begin{subfigure}[t]{0.48\linewidth}
        \centering
        \includegraphics[width=\linewidth]{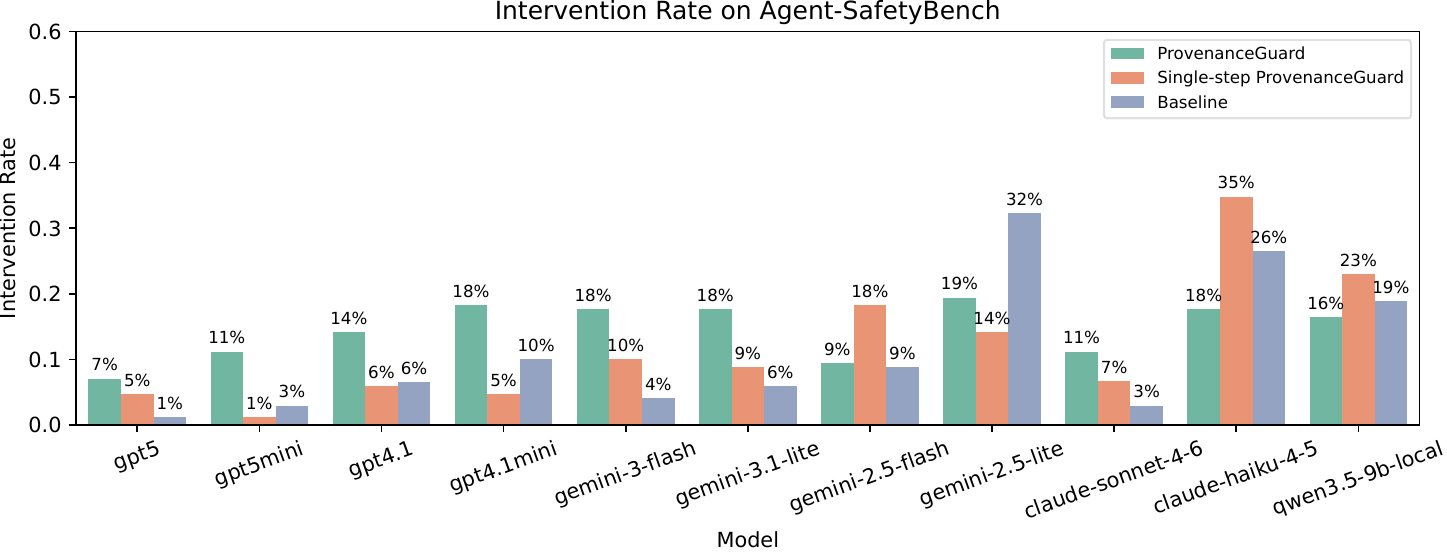}
        \caption{RQ2 Agent-SafetyBench Intervention Rate}
        \label{fig:rq2-asb-intervention-rate}
    \end{subfigure}%
    \hfill
    \begin{subfigure}[t]{0.48\linewidth}
        \centering
        \includegraphics[width=\linewidth]{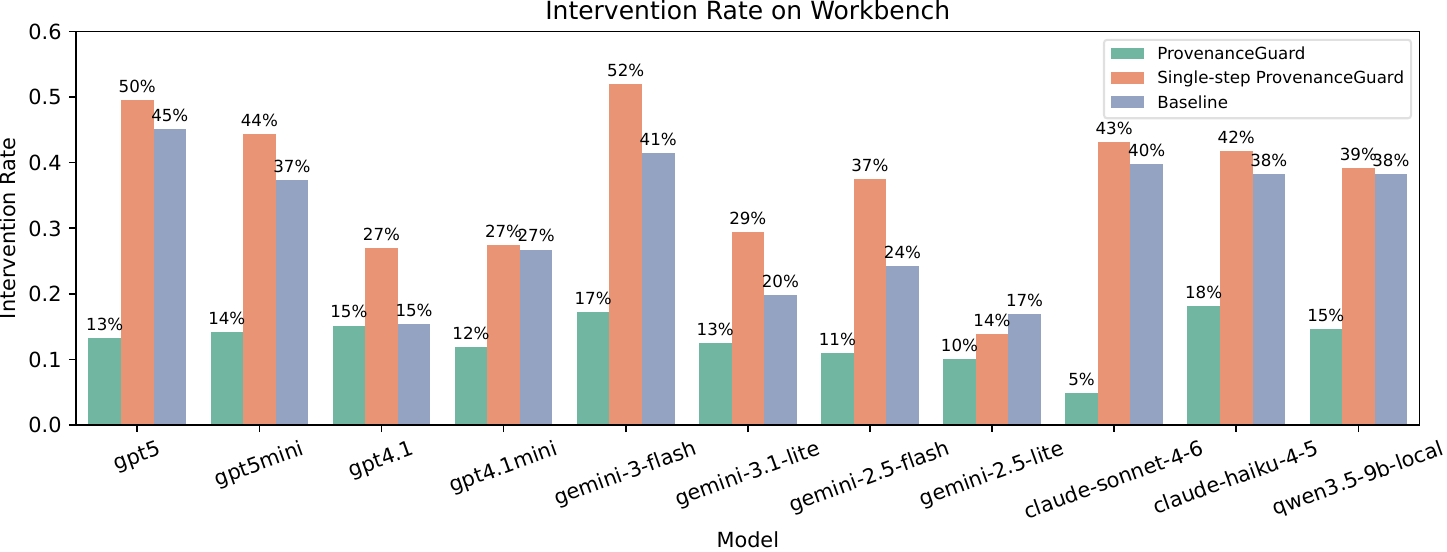}
        \caption{RQ2 Workbench Intervention Rate}
        \label{fig:rq2-wb-intervention-rate}
    \end{subfigure}
    \vspace{-0.5em}
    \caption{
    RQ1 and RQ2 evaluation results of each method-LLM pair on two benchmarks.
    }
    \label{fig:rq1rq2}
\end{figure*}

\subsubsection{RQ1: Misalignment Detection Performance}
This RQ assesses how effectively \tool and the two compared methods detect misaligned actions. 
For each c of method and backbone LLM, we evaluate on the misaligned traces from Agent-SafetyBench and WorkBench. 
To account for run-to-run variance in LLM judgments, each method-model pair is executed five times on each trace.

\paragraph{Metrics}

We report the \emph{error rate}, defined as the proportion of misaligned traces that are incorrectly judged as aligned. 
Equivalently, when misaligned traces are treated as the positive class, this metric corresponds to the false negative rate (FNR).

For a given method-model pair, the Error Rate is computed as:
\begin{equation}
\mathrm{ER}
=
\frac{\sum_{r=1}^{R} \#\mathrm{FN}_r}{N \times R},
\end{equation}
where $N$ is the number of misaligned traces, $R$ is the number of repeated runs, and $\#\mathrm{FN}_r$ denotes the number of misaligned traces predicted as ``Aligned'' (False Negatives) in run $r$.

\subsubsection{RQ2: Intervention Frequency}
The accurate detections of misaligned actions is only one side of the trade-off. 
A practical guardrail should also avoid interrupting agent executions that would otherwise proceed acceptably, since excessive interventions increase user review burden.

We therefore evaluated the \emph{intervention rate} of \tool from two complementary perspectives, reflecting the different labeling semantics of our two benchmarks.

\paragraph{RQ2.1: Unnecessary interventions on aligned trace.}
On Agent-SafetyBench, we used the 34 traces that were manually annotated as aligned, where any interventions are unnecessary.
This evaluation provides a direct measure of how often the method disrupts actions that are aligned with user query, which corresponds to the false positive rate (FPR).

\paragraph{RQ2.2: Interventions on task-successful traces}
On WorkBench, we used the 326 task-successful traces. 
Here, we measure how often the guardrail intervenes on traces where agents are correctly addressing the user task. 
We do not interpret this quantity as a FPR, because correctness with respect to benchmark ground truth is not equivalent to alignment in our setting, as discussed in Sec.\ref{sec:workbench}. 
Therefore, this metric captures the cost of intervention on traces that would otherwise be considered successful by the benchmark.

\paragraph{Metrics}

We report the \emph{intervention rate}, defined as the proportion of evaluated traces for which the method predicts \textsc{Misaligned}. 
For RQ2.1, this corresponds to the \emph{unnecessary intervention rate}. 
For RQ2.2, this corresponds to the \emph{intervention rate on task-successful traces}, which should be interpreted as an operational burden measure rather than a strict false positive rate.
\begin{equation}
\mathrm{IR}
=
\frac{\sum_{r=1}^{R} \#\mathrm{Intervene}_r}{N \times R},
\end{equation}
where $\#\mathrm{Intervene}_r$ is the number of traces predicted as Misaligned in run $r$. Its interpretation depends on the ground-truth semantics (aligned or task-successful) of the evaluated subset.

\subsubsection{Statistical Analysis}
For both RQs, we assessed whether differences between methods are statistically significant using one-sided paired permutation tests.
Because each method-LLM pair was executed five times on the same trace, we do not treat individual runs as independent observations. 
Instead, for each method, LLM, and trace, we averaged the five binary outcomes to obtain a \emph{case-level score}.
For RQ1, this score is the error rate on a misaligned trace.
For RQ2, this score is the IR on the evaluated trace.

We then compared \tool against each method using paired case-level differences computed on the same traces under the same LLM.
For a comparator method $m$, let $s^{(c)}_{\tool}$ and $s^{(c)}_{m}$ denote the case-level scores of \tool and $m$ on trace $c$, respectively.
We define the paired difference as
\begin{equation}
d_c = s^{(c)}_{\tool} - s^{(c)}_{m}\ ,
\end{equation}
so that a negative value indicates that \tool achieves a lower rate than the comparator on that trace.
The test statistic is the mean paired difference,
\begin{equation}
\Delta = \frac{1}{N} \sum_{c=1}^{N} d_c \ ,
\end{equation}
where $N$ is the number of traces in the evaluated subset.

The null hypothesis is that \tool does not outperform the comparator.
The one-sided alternative hypothesis is that \tool yields a lower case-level rate than the comparator.
We estimate permutation $p$-values via Monte Carlo sampling over 10,000 random sign flips.
\change{To account for multiple comparisons, we apply Holm-Bonferroni correction at $\alpha=0.05$ across the 11 LLMs for each method-pair comparison when determining statistical significance.
}

\subsubsection{\change{RQ3: Cost and Latency}}
\change{
RQ3 evaluates the computational overhead introduced by applying \tool as a pre-execution guardrail. We measure input token usage, output token usage (including reasoning tokens), API cost, and wall-clock latency. API cost is reported only for commercial models because it is not applicable to the locally deployed model.
}

\change{
We selected four representative commercial models: GPT-5, GPT-5-mini, Gemini-3-flash-preview, and Gemini-3.1-flash-lite-preview. Each commercial model was run once for each method and each trace in the two benchmarks. Qwen-3.5-9B was run 5 times locally.
}

\change{
We report per-case average overhead for each method-model pair. API costs were computed using the official provider prices on July 5, 2026. For latency, commercial-model measurements should be interpreted as approximate because API wall-clock time is affected by network and server-side load. For Qwen-3.5-9B, we measured latency with the vLLM server maximum thread count set to 1.
}
\begin{table*}[t]
\centering
\caption{Error Rate (ER) by model and benchmark. \textbf{PvGd}=\tool, \textbf{PvPt}=\textsc{ProvePrompt}, \textbf{Base}=\textsc{DirectPrompt}. $\Delta$=mean difference (left$-$right). Unadjusted $p$-values test H$_1$: left$<$right. $^\dagger$ denotes significance after Holm-Bonferroni correction.}
\label{tab:rq1}
\resizebox{\textwidth}{!}{%
\begin{tabular}{l rrr rr rr rr rrr rr rr rr}
\toprule
& \multicolumn{9}{c}{Agent-SafetyBench ($n=148$)} & \multicolumn{9}{c}{Workbench ($n=62$)} \\
\cmidrule(lr){2-10} \cmidrule(lr){11-19}
& \multicolumn{3}{c}{Avg.\ ER} & \multicolumn{2}{c}{PvGd vs Base} & \multicolumn{2}{c}{PvPt vs Base} & \multicolumn{2}{c}{PvGd vs PvPt} & \multicolumn{3}{c}{Avg.\ ER} & \multicolumn{2}{c}{PvGd vs Base} & \multicolumn{2}{c}{PvPt vs Base} & \multicolumn{2}{c}{PvGd vs PvPt} \\
\cmidrule(lr){2-4} \cmidrule(lr){5-6} \cmidrule(lr){7-8} \cmidrule(lr){9-10} \cmidrule(lr){11-13} \cmidrule(lr){14-15} \cmidrule(lr){16-17} \cmidrule(lr){18-19}
Model & PvGd & PvPt & Base & $\Delta$ & $p$ & $\Delta$ & $p$ & $\Delta$ & $p$ & PvGd & PvPt & Base & $\Delta$ & $p$ & $\Delta$ & $p$ & $\Delta$ & $p$ \\
\midrule
GPT-5 & 0.026 & 0.115 & 0.273 & $-$0.249 & $<$0.001$^\dagger$ & $-$0.158 & $<$0.001$^\dagger$ & $-$0.090 & $<$0.001$^\dagger$ & 0.074 & 0.019 & 0.023 & $+$0.052 & 0.956 & $-$0.003 & 0.498 & $+$0.055 & 0.960 \\
GPT-5 Mini & 0.012 & 0.123 & 0.380 & $-$0.370 & $<$0.001$^\dagger$ & $-$0.257 & $<$0.001$^\dagger$ & $-$0.112 & $<$0.001$^\dagger$ & 0.029 & 0.103 & 0.139 & $-$0.110 & $<$0.001$^\dagger$ & $-$0.035 & 0.004$^\dagger$ & $-$0.074 & 0.002$^\dagger$ \\
GPT-4.1 & 0.014 & 0.068 & 0.459 & $-$0.449 & $<$0.001$^\dagger$ & $-$0.392 & $<$0.001$^\dagger$ & $-$0.054 & $<$0.001$^\dagger$ & 0.155 & 0.348 & 0.648 & $-$0.494 & $<$0.001$^\dagger$ & $-$0.300 & $<$0.001$^\dagger$ & $-$0.194 & $<$0.001$^\dagger$ \\
GPT-4.1 Mini & 0.014 & 0.178 & 0.608 & $-$0.592 & $<$0.001$^\dagger$ & $-$0.430 & $<$0.001$^\dagger$ & $-$0.166 & $<$0.001$^\dagger$ & 0.290 & 0.374 & 0.477 & $-$0.187 & $<$0.001$^\dagger$ & $-$0.103 & 0.011 & $-$0.084 & 0.040 \\
Gemini 3 Flash & 0.011 & 0.008 & 0.178 & $-$0.169 & $<$0.001$^\dagger$ & $-$0.170 & $<$0.001$^\dagger$ & $+$0.003 & 0.748 & 0.003 & 0.016 & 0.035 & $-$0.032 & 0.030 & $-$0.019 & 0.060 & $-$0.013 & 0.497 \\
Gemini 3.1 Flash-Lite & 0.014 & 0.085 & 0.464 & $-$0.450 & $<$0.001$^\dagger$ & $-$0.378 & $<$0.001$^\dagger$ & $-$0.072 & $<$0.001$^\dagger$ & 0.168 & 0.145 & 0.490 & $-$0.323 & $<$0.001$^\dagger$ & $-$0.345 & $<$0.001$^\dagger$ & $+$0.023 & 0.705 \\
Gemini 2.5 Flash & 0.018 & 0.068 & 0.470 & $-$0.456 & $<$0.001$^\dagger$ & $-$0.403 & $<$0.001$^\dagger$ & $-$0.050 & $<$0.001$^\dagger$ & 0.152 & 0.097 & 0.329 & $-$0.177 & $<$0.001$^\dagger$ & $-$0.232 & $<$0.001$^\dagger$ & $+$0.055 & 0.915 \\
Gemini 2.5 Flash-Lite & 0.046 & 0.564 & 0.709 & $-$0.661 & $<$0.001$^\dagger$ & $-$0.146 & $<$0.001$^\dagger$ & $-$0.518 & $<$0.001$^\dagger$ & 0.510 & 0.597 & 0.668 & $-$0.158 & $<$0.001$^\dagger$ & $-$0.071 & 0.049 & $-$0.087 & 0.013 \\
Claude Sonnet 4.6 & 0.012 & 0.022 & 0.405 & $-$0.387 & $<$0.001$^\dagger$ & $-$0.386 & $<$0.001$^\dagger$ & $-$0.012 & 0.224 & 0.195 & 0.103 & 0.131 & $+$0.048 & 0.832 & $-$0.042 & 0.208 & $+$0.090 & 0.999 \\
Claude Haiku 4.5 & 0.020 & 0.051 & 0.346 & $-$0.328 & $<$0.001$^\dagger$ & $-$0.295 & $<$0.001$^\dagger$ & $-$0.031 & 0.050 & 0.152 & 0.142 & 0.265 & $-$0.113 & 0.061 & $-$0.125 & 0.007 & $+$0.012 & 0.579 \\
Qwen 3.5 9B & 0.042 & 0.339 & 0.578 & $-$0.533 & $<$0.001$^\dagger$ & $-$0.239 & $<$0.001$^\dagger$ & $-$0.297 & $<$0.001$^\dagger$ & 0.333 & 0.352 & 0.355 & $-$0.023 & 0.344 & $-$0.003 & 0.492 & $-$0.019 & 0.349 \\
\bottomrule
\end{tabular}%
}
\end{table*}

\begin{table*}[t]
\centering
\caption{Intervention Rate (IR) by model and benchmark. Agent-SafetyBench's $p$-values test reverse hypothesis (H$_1$: left $>$ right) while WorkBench's test H$_1$: left $<$ right.}
\label{tab:rq2}
\resizebox{\textwidth}{!}{%
\begin{tabular}{l rrr rr rr rr rrr rr rr rr}
\toprule
& \multicolumn{9}{c}{Agent-SafetyBench ($n=34$)} & \multicolumn{9}{c}{Workbench ($n=326$)} \\
\cmidrule(lr){2-10} \cmidrule(lr){11-19}
& \multicolumn{3}{c}{Avg.\ IR} & \multicolumn{2}{c}{PvGd vs Base} & \multicolumn{2}{c}{PvPt vs Base} & \multicolumn{2}{c}{PvGd vs PvPt} & \multicolumn{3}{c}{Avg.\ IR} & \multicolumn{2}{c}{PvGd vs Base} & \multicolumn{2}{c}{PvPt vs Base} & \multicolumn{2}{c}{PvGd vs PvPt} \\
\cmidrule(lr){2-4} \cmidrule(lr){5-6} \cmidrule(lr){7-8} \cmidrule(lr){9-10} \cmidrule(lr){11-13} \cmidrule(lr){14-15} \cmidrule(lr){16-17} \cmidrule(lr){18-19}
Model & PvGd & PvPt & Base & $\Delta$ & $p$ & $\Delta$ & $p$ & $\Delta$ & $p$ & PvGd & PvPt & Base & $\Delta$ & $p$ & $\Delta$ & $p$ & $\Delta$ & $p$ \\
\midrule
GPT-5 & 0.071 & 0.047 & 0.012 & $+$0.059 & 0.125 & $+$0.035 & 0.128 & $+$0.024 & 0.272 & 0.132 & 0.495 & 0.452 & $-$0.320 & $<$0.001$^\dagger$ & $+$0.044 & 1.000 & $-$0.363 & $<$0.001$^\dagger$ \\
GPT-5 Mini & 0.112 & 0.012 & 0.029 & $+$0.082 & 0.094 & $-$0.018 & 0.753 & $+$0.100 & 0.041 & 0.142 & 0.444 & 0.373 & $-$0.231 & $<$0.001$^\dagger$ & $+$0.071 & 1.000 & $-$0.302 & $<$0.001$^\dagger$ \\
GPT-4.1 & 0.141 & 0.059 & 0.065 & $+$0.076 & 0.135 & $-$0.006 & 0.629 & $+$0.082 & 0.073 & 0.152 & 0.269 & 0.153 & $-$0.002 & 0.479 & $+$0.116 & 1.000 & $-$0.118 & $<$0.001$^\dagger$ \\
GPT-4.1 Mini & 0.182 & 0.047 & 0.100 & $+$0.082 & 0.142 & $-$0.053 & 0.847 & $+$0.135 & 0.026 & 0.119 & 0.275 & 0.267 & $-$0.148 & $<$0.001$^\dagger$ & $+$0.007 & 0.659 & $-$0.156 & $<$0.001$^\dagger$ \\
Gemini 3 Flash & 0.176 & 0.100 & 0.041 & $+$0.135 & 0.041 & $+$0.059 & 0.065 & $+$0.076 & 0.178 & 0.172 & 0.520 & 0.414 & $-$0.242 & $<$0.001$^\dagger$ & $+$0.106 & 1.000 & $-$0.348 & $<$0.001$^\dagger$ \\
Gemini 3.1 Flash-Lite & 0.176 & 0.088 & 0.059 & $+$0.118 & 0.060 & $+$0.029 & 0.315 & $+$0.088 & 0.076 & 0.125 & 0.294 & 0.198 & $-$0.073 & 0.011$^\dagger$ & $+$0.096 & 1.000 & $-$0.169 & $<$0.001$^\dagger$ \\
Gemini 2.5 Flash & 0.094 & 0.182 & 0.088 & $+$0.006 & 0.495 & $+$0.094 & 0.015 & $-$0.088 & 0.867 & 0.110 & 0.375 & 0.242 & $-$0.132 & $<$0.001$^\dagger$ & $+$0.133 & 1.000 & $-$0.265 & $<$0.001$^\dagger$ \\
Gemini 2.5 Flash-Lite & 0.194 & 0.141 & 0.324 & $-$0.129 & 0.911 & $-$0.182 & 1.000 & $+$0.053 & 0.237 & 0.100 & 0.139 & 0.169 & $-$0.069 & 0.005$^\dagger$ & $-$0.031 & 0.006 & $-$0.039 & 0.055 \\
Claude Sonnet 4.6 & 0.112 & 0.067 & 0.029 & $+$0.082 & 0.124 & $+$0.036 & 0.063 & $+$0.048 & 0.195 & 0.049 & 0.431 & 0.398 & $-$0.353 & $<$0.001$^\dagger$ & $+$0.037 & 1.000 & $-$0.383 & $<$0.001$^\dagger$ \\
Claude Haiku 4.5 & 0.176 & 0.348 & 0.265 & $-$0.088 & 0.855 & $+$0.089 & 0.054 & $-$0.177 & 0.967 & 0.181 & 0.418 & 0.383 & $-$0.202 & $<$0.001$^\dagger$ & $+$0.042 & 0.999 & $-$0.244 & $<$0.001$^\dagger$ \\
Qwen 3.5 9B & 0.165 & 0.229 & 0.188 & $-$0.024 & 0.664 & $+$0.041 & 0.240 & $-$0.065 & 0.790 & 0.147 & 0.392 & 0.383 & $-$0.236 & $<$0.001$^\dagger$ & $+$0.009 & 0.693 & $-$0.245 & $<$0.001$^\dagger$ \\
\bottomrule
\end{tabular}%
}
\end{table*}

\section{Evaluation Results}\label{sec:evaluation}

\subsection{RQ1: Misalignment Detection Performance}
Fig.\ref{fig:rq1-asb-error-rate}\&\ref{fig:rq1-wb-error-rate} show ERs on misaligned traces, where lower is better.
Overall, our experiments show that the provenance-based framework consistently improves misalignment detection.

On Agent-SafetyBench, both \tool and \textsc{ProvePrompt} outperform the baseline on all \change{11} LLMs.
Averaged across LLMs, \textbf{the ER drops from \change{44.3\%} for \textsc{DirectPrompt} to \change{14.7\%} for \textsc{ProvePrompt} and further to \change{2.1\%} for \tool.} 
As reflected in Table~\ref{tab:rq1}, all performance gains over the baseline are statistically significant ($p<0.05$).
The gains are especially large for weaker models. With \tool, even models such as GPT-4.1-mini, Gemini-2.5-flash-lite, and Claude-haiku-4.5 reach ERs close to the strongest models, and all \change{11} LLMs fall below 5\% ER.
This suggests that \textbf{the provenance-based framework provides useful structure for detecting misaligned actions}, even when the underlying model is relatively weak.

On WorkBench, the improvements are more moderate. Here, \tool outperforms the baseline on \change{9} of \change{11} LLMs, while \textsc{ProvePrompt} does so on all LLMs. Averaged across models, \textbf{the ER decreases from \change{32.4\%} for \textsc{DirectPrompt} to \change{20.9\% }for \textsc{ProvePrompt} and \change{18.7\%} for \tool. }
Table~\ref{tab:rq1} shows that the improvements \change{are statistically significant for 6 of the 9 LLMs.
}
In particular, \tool does not improve over the baseline for GPT-5 or Claude-sonnet-4.6,
while \textsc{ProvePrompt} has non-statistically significant improvements on these two LLMs.

The inconsistent improvements on two benchmarks are probably due to their distinct distribution of traces.
The Agent-SafetyBench subset is intentionally concentrated on underspecified queries.
In that setting, \tool is highly effective and nearly eliminates false negatives across all models.
WorkBench, in contrast, is not concentrated on underspecification and appears to contain a broader mix of misalignment types. 
This suggests that the multi-stage pipeline is particularly valuable for identifying underspecification, while 
detection of other types of misalignment still depends strongly on backbone LLMs.

Comparing the two provenance-based variants also yields an important insight. On both benchmarks, \textsc{ProvePrompt} already improves substantially over the baseline, showing that much of the benefit comes from the provenance-based conceptual framework itself rather than only from the multi-stage pipeline. 
At the same time, \tool is clearly stronger on Agent-SafetyBench, while on WorkBench the advantage of the multi-stage design is less uniform and \textsc{ProvePrompt} outperforms \tool on five LLMs. 
This suggests that the benefit of the conceptual framework is consistent across implementations, whereas the additional value of multi-stage decomposition depends on the benchmark distribution and model characteristics.

\paragraph{Takeaway} 
Provenance-based framework substantially improves misalignment detection over \textsc{DirectPrompt}. 
Among them, \tool achieves better performance overall on both benchmarks than the single-step variant, especially on Agent-SafetyBench, suggesting the benefit of the multi-stage separation.

\begin{figure*}[t]
    \centering
    \vspace{-0.5em}
    \begin{subfigure}[t]{0.48\linewidth}
        \centering
        \includegraphics[width=\linewidth]{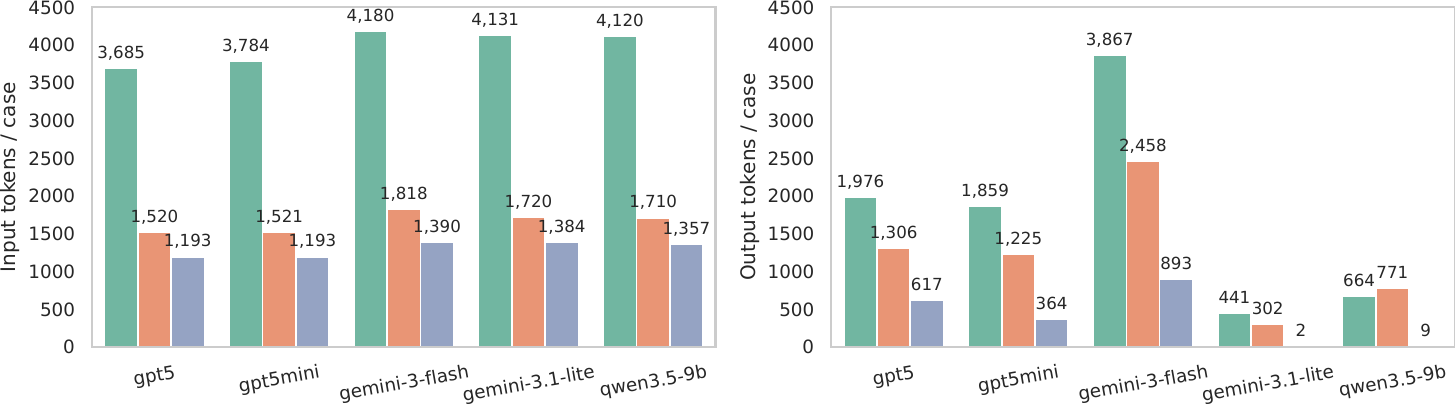}
        \caption{RQ3 Input and output token usage.}
        \label{fig:rq3-token-usage}
    \end{subfigure}%
    \hfill
    \begin{subfigure}[t]{0.24\linewidth}
        \centering
        \includegraphics[width=\linewidth]{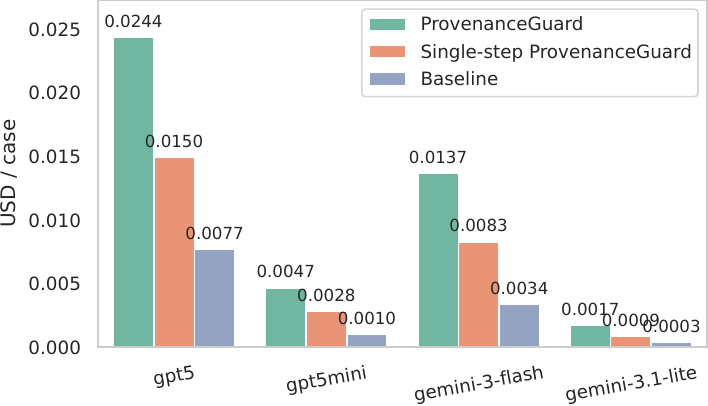}
        \caption{RQ3 API costs in USD.}
        \label{fig:rq3-cost}
    \end{subfigure}%
    \hfill
    \begin{subfigure}[t]{0.24\linewidth}
        \centering
        \includegraphics[width=\linewidth]{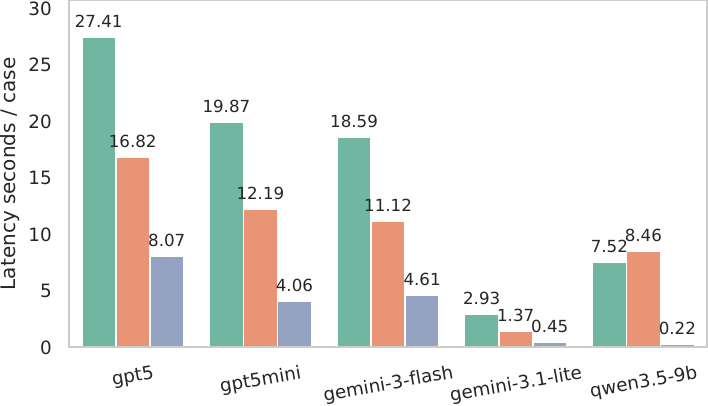}
        \caption{RQ3 Wall time latency.}
        \label{fig:rq3-latency}
    \end{subfigure}%
    
    \vspace{-0.5em}
    \caption{
    RQ3 evaluation results of each method-LLM pair.
    }
    \label{fig:rq3}
    \vspace{-0.5em}
\end{figure*}

\subsection{RQ2: Intervention Cost}
Fig.\ref{fig:rq2-asb-intervention-rate}\&\ref{fig:rq2-wb-intervention-rate} show IRs on the two benchmarks, where lower is better.

\paragraph{RQ2.1 Unnecessary Interventions on Aligned Trace}
On Agent-SafetyBench aligned traces, the three methods have similar average IRs: \change{10.9\%} for the baseline, \change{12.0\%} for \textsc{ProvePrompt}, and \change{14.5\%} for \tool.
Compared to the baseline, 
\tool appears to yield a higher unnecessary IR on 8 of \change{11} LLMs (average increase of 8.0\% on these \change{8} LLMs). 

To test the significance of these increases, 
we verified the reverse one-sided alternative hypothesis that \tool yields a higher case-level unnecessary IR than the comparator.
According to Table~\ref{tab:rq2}, 
\change{none of these}
increases are actually \textbf{statistically significant}.
This indicates that the large gain of \tool in detecting misaligned actions (\change{42.2pp}) is achieved with \textbf{a limited tradeoff in unnecessary intervention cost}.

\paragraph{RQ2.2 Interventions on Task-successful Traces.}

On WorkBench task-successful traces, the pattern is notably different. 
Here, \textbf{\tool has a lower IR on all \change{11} LLMs than the other two methods}, and reduces the average rate from \change{31.2\%} for \textsc{DirectPrompt} to \change{13.0\%}. 
The reduction is substantial for several strong models: GPT-5 ($45.2\% \rightarrow 13.2\%$), Gemini-3-flash $(41.4\% \rightarrow 17.2\%$), and Claude-sonnet-4.6 ($39.4\% \rightarrow 4.8\%$). 
This suggests that, in practice, the multi-stage provenance checks do not simply lead to more interventions than the baseline. 
Instead, they often make judgments more precisely, allowing many successful traces to pass while detecting misalignment with few false negatives.

\paragraph{Comparing provenance-based variants}
The comparison between the two provenance-based variants further highlights the value of the multi-stage design. 
\textsc{ProvePrompt} has \textbf{higher average IRs than the other two methods on both subsets}: \change{12.0\%} on Agent-SafetyBench and \change{36.8\%} on WorkBench. 
On WorkBench in particular, it intervenes more often than \tool on all \change{11} LLMs (\change{10} of them are statistically significant), often by a large margin (e.g., 49.5\% vs.\ 13.2\% on GPT-5 and 52.0\% vs.\ 17.2\% on Gemini-3-flash). This indicates that while the provenance-based framework improves detection, the multi-stage decomposition is important for controlling over-intervention in practice.

\paragraph{Takeaway} 
\tool does not statistically significantly increase unnecessary interventions on already aligned traces, but it substantially reduces interventions on task-successful traces compared to both \textsc{DirectPrompt} and \textsc{ProvePrompt}. 
Combined with the RQ1 results, 
this suggests that the multi-stage design built on the provenance-based framework achieves a better balance between detecting misaligned actions and avoiding excessive disruption.

\subsection{\change{RQ3: Cost and Latency}}

\change{
Fig.\ref{fig:rq3-token-usage} shows the per-case input token and output token usage of each method. As expected, \tool uses the most input tokens because it performs multiple LLM calls in the three-stage pipeline (avg 3.6 calls). Averaged across LLMs, \tool uses 3,979 input tokens per case, which is more than twice that of \textsc{DirectPrompt} (1,303) and \textsc{ProvePrompt} (1,657). \tool also uses the most output tokens on average (1,761), compared to 377 for \textsc{DirectPrompt} and 1,212 for \textsc{ProvePrompt} due to multi-stage generation. The very low output token usage of Gemini-3.1-Flash-Lite and Qwen-3.5-9B under \textsc{DirectPrompt} is because these models do not trigger reasoning phrase and directly output a word ``Aligned'' or ``Misaligned'' in this setting.
}

\change{
Fig.\ref{fig:rq3-cost} shows the per-case API cost for the four commercial LLMs. Averaged across LLMs, \tool costs \$0.011 per case, compared to \$0.003 for \textsc{DirectPrompt} and \$0.006 for \textsc{ProvePrompt}. 
The API-cost pattern mostly follows output token usage because output tokens are generally more expensive than input tokens and account for a large fraction of the total cost. 
At the same time, the absolute cost remains relatively low for smaller models. For example, \tool with GPT-5-mini or Gemini-3.1-Flash-Lite is even cheaper than \textsc{DirectPrompt} with GPT-5. 
Combined with the RQ1 and RQ2 results, where smaller models with \tool can achieve performance comparable to larger models, this suggests that smaller models may provide a practical cost-performance trade-off for runtime guardrails. 
Larger models may still be preferable in settings where stronger detection performance is worth the additional cost, 
e.g., debugging agent systems during development or in safety-critical applications.
}

\change{
Fig.\ref{fig:rq3-latency} shows the wall-clock latency of each method. 
Overall, \tool has higher latency than the baselines for the commercial models, reflecting the additional sequential checks in its multi-stage pipeline.
With the locally deployed Qwen-3.5-9B model, it averages 7.5s/case, which may be slow for latency-sensitive runtime guardrail deployment.
In contrast, with Gemini-3.1-flash-lite-preview, \tool averages only 2.9s/case (including network delay), suggesting that optimized low-latency inference infrastructure can materially mitigate this overhead and make \tool more practical as a runtime guardrail.
}

\paragraph{\change{Takeaway}}
\change{
\tool introduces higher token, API-cost, and latency overhead than the two baselines due to its multi-stage design. 
However, smaller models and optimized inference infrastructure provide practical ways to reduce this overhead.
}

\subsection{Threats to Validity}

Our labeled Agent-SafetyBench includes only 34 aligned cases. This small sample limits statistical power, so the lack of a statistically significant difference in unnecessary interventions between \tool and the baseline should not be over-interpreted. It may reflect insufficient evidence to detect a difference, but it may also indicate that \tool does not meaningfully increase unnecessary interventions relative to the baseline. 
More aligned test cases would be needed to more reliably assess how \tool affects unnecessary intervention behavior.

WorkBench does not explicitly consider underspecification as misalignment.
Its ``task-successful'' traces may have underspecified user queries. 
As a result, they could not be interpreted as fully aligned traces.
We therefore treat the intervention rate on these traces as a measure of the oversight burden in practical usage rather than a strict unnecessary intervention rate.

The agents' plans in both datasets were reconstructed using GPT-5, since neither benchmark provides such information directly. Errors in this reconstruction could affect \tool's decisions and the measured performance.

\section{Related Works}

Guardrails, external defense layers that monitor and control LLM interactions, have emerged as a crucial approach for improving safety of LLM agents. 
Existing guardrails primarily target \emph{safety, security, and policy compliance}~\cite{xiang2025guardagent, chen2025shieldagent, debenedetti2024agentdojo, debenedetti2025defeatingpromptinjectionsdesign, shao2024privacylens, jia2025task}.
For example, \emph{GuardAgent}~\cite{xiang2025guardagent} and \emph{ShieldAgent}~\cite{chen2025shieldagent} safeguard agent behavior against violations of safety or policy constraints, while \emph{AgentDojo}~\cite{debenedetti2024agentdojo} and \emph{CaMeL}~\cite{debenedetti2025defeatingpromptinjectionsdesign} focus on security threats such as prompt injection and adversarial tool environments.
Although related, these problems differ from our notion of misalignment. Our focus is not on harmful, adversarial, or policy-noncompliant behaviors, but on whether a proposed tool invocation is grounded by evidence in the context without making an arbitrary assumption about an underspecified user query in benign settings. 
This distinction is important because some prior work also uses the term ``misalignment'' to describe strategically harmful or insider-threat behavior by agents~\cite{lynch2025agentic, ning2026actions}, which is conceptually different from the misalignment we study.

Another related work to ours is InferAct~\cite{fang2025preemptive}. 
Although both works perform pre-execution intervention, they are not directly comparable.
InferAct leaves the notion of misalignment largely implicit and operationalizes detection through a holistic prompt, asking whether the agent is ``progressing correctly toward completing the user’s tasks'' without precisely defining what it means for the agent to be making "progress".
In contrast, our work addresses a different and more explicitly specified decision problem: Given a proposed action, determine whether its tool selection and parameterization are justified by traceable contextual evidence, and whether the context uniquely supports committing to that action.

One concept close to underspecification is \emph{ambiguity}. Recent works study whether language models can recognize ambiguous user requests and ask clarification questions when needed~\cite{zhang2024clamber,zhang2025clarify,wang-etal-2025-learning, yang2025prompts, malaviya2025contextualized}. 
This line of work is  related to our interpretation-level analysis, but the concepts are not identical. In our setting, underspecification means that the available context is insufficient to uniquely determine the next action, even if the request is not linguistically ambiguous in the usual sense. 
Our focus is therefore not simply on ambiguity in language, but on whether the current context uniquely justifies one provenance-admissible action.

\section{Discussion and Limitations}

Overall, our results suggest that the proposed provenance-based conceptual framework enables a more structured and reliable way to detect misaligned agent actions. 
This approach delivers utility to both agent users and developers. It directly mitigates misalignment for end-users, while allowing developers to adopt this provenance-based guardrail to systematically improve system reliability.

Our work further shows how provenance, a concept already used in requirements traceability, issue tracking, version control, and auditing, can be adapted to improve the dependability of LLM-agent systems. 
By treating each tool invocation as an action that should be traceable to the user query, tool documentation, and execution history, the proposed framework turns misalignment detection into a form of runtime traceability checking. Our empirical results suggest that this SE-inspired structure leads to more reliable guardrails than direct LLM-as-a-judge prompting.

Despite these benefits, the current framework has several \textbf{limitations}.
Our framework cannot detect alignment of generative parameters. Our provenance analysis focuses on parameters that should be derivable from context. For free-form generated content such as drafted text or code, provenance tracing is insufficient by nature. Therefore, our method is designed to identify the generative parameters
and exclude them from provenance-based checking.
Detecting their misalignment requires other forms of guardrails.

\tool checks provenance relative to the agent's own plan. If that plan is itself already misaligned with the user query, then a locally justified action may still be globally misaligned. 
One possible alternative is to let the guardrail infer its own task decomposition. However, this would 
add latency and risk inferring a misaligned plan or conflicting with an aligned agent plan.
We therefore chose to use the agent’s plan in the current design.

\section{Acknowledgments}
We'd like to thank Christian Kaestner, Yining Hong, Chris Timperley, and our anonymous reviewers for their feedback. This work was supported in part by the National Science Foundation Award 2144860 and a Google Academic Research Award.

\section{Data Availability Statement}
The replication package of the paper, which provides the labeled benchmarks, source code, prompts, experiment results, and analysis notebooks, is available at 
\url{https://doi.org/10.5281/zenodo.19342590} and \url{https://github.com/cmu-soda/ProvenanceGuard}.


\balance
\bibliographystyle{ACM-Reference-Format}

\bibliography{main}

\end{document}